# Current Studies and Applications of Shuffled Frog Leaping Algorithm: A Review

Bestan B. Maaroof[1]; Tarik A. Rashid[2]; Jaza M. Abdulla[3,4]; Bryar A. Hassan[5]; Abeer Alsadoon[6,7]; Mokhtar Mohammadi[8]; Mohammad Khishe[9]; Seyedali Mirjalili[10,11]

[1]Information Technology Department, College of Commerce, University of Sulaimani, KRI, Iraq

[2]Computer Science and Engineering, University of Kurdistan Hewler, Erbil, KRI, Iraq.

[3]Computer Science Department, College of Science, Komar University of Science and Technology, Sulaymaniyah, KRI, Iraq.

[4]Information Technology, College of Commerce, University of Sulaimani, Sulaymaniyah, KRI, Iraq.

[5]Kurdistan Institution for Strategic Studies and Scientific Research, Sulaimani, Iraq

[6]School of Computing and Mathematics, Charles Sturt University, Sydney, Australia

[7]Asia Pacific International College (APIC), Information Technology Department, Sydney, Australia

[8]Department of Information Technology, College of Engineering and Computer Science, Lebanese French University, Kurdistan Region, Iraq

[9]Department of Marine Electronics and Communication Engineering, Imam Khomeini Marine Science University, Nowshahr, Iran

[10]Centre for Artificial Intelligence Research and Optimization, Torrens University, Australia

[11]Yonsei Frontier Lab, Yonsei University, Seoul, Korea

Emails (corresponding): bryar.hassan@kissr.edu.krd; tarik.ahmed@ukh.edu.krd

**Abstract**

Shuffled Frog Leaping Algorithm (SFLA) is one of the most widespread algorithms. It was developed by Eusuff and Lansey in 2006. SFLA is a population-based metaheuristic algorithm that combines the benefits of memetics with particle swarm optimization. It has been used in various areas, especially in engineering problems due to its implementation easiness and limited variables. Many improvements have been made to the algorithm to alleviate its drawbacks, whether they were achieved through modifications or hybridizations with other well-known algorithms. This paper reviews the most relevant works on this algorithm. An overview of the SFLA is first conducted, followed by the algorithm's most recent modifications and hybridizations. Next, recent applications of the algorithm are discussed. Then, an operational framework of SLFA and its variants is proposed to analyze their uses on different cohorts of applications. Finally, future improvements to the algorithm are suggested. The main incentive to conduct this survey to provide useful information about the SFLA to researchers interested in working on the algorithm's enhancement or application.

**Keywords**

Shuffled Leap-Frog Algorithm (SFLA), Hybridization, Modification Application, Review.

## 1. Introduction

The SFLA is originated as a result of an investigation by Eusuf and Lansey in 2003. They combined the ideas used in the Shuffled Complex Evolution algorithm (SCE) [1] and Particle Swarm Optimization (PSO) [2]. Consequently, they come up with a new meta-heuristic for tackling discrete and/or combinatorial problems. The basic architecture of the SFLA is based on the assumptions that knowledge is shared socially between the population, providing an evolutionary benefit [3].

SFLA is a methodology, which depends on imitation of the behavior patterns of frogs, taking into account a crowd of frogs leaping in a swamp, on the lookout for the place, which has the highest food quantity reachable, in which the swamp has multiple stones at distinct points that make it easier for the frogs to step on. The aim is to identify a stone with the maximum food amount available. Communicating between frogs can progress their memes as infection can be propagated among them. As a result of improvement in memes, each frog's position will be changed by tuning its leaping step size [4]. The population is meant to be a number of solutions which is represented as a swarm of frogs that is segregated into subgroups denoted to as memeplexes. Each memeplex represented a distinct frog community, each conducting a local quest. The specific frogs grip ideas within each memeplex, which can be persuaded by further frogs' ideas and progress through a memetic evolution process. In a shuffling process, ideas are passed between memeplexes exactly after several memetic evolution phases [5].

This algorithm's main advantages are its fast convergence and accuracy in searching for global solutions [6]. The first SFLA proposal was implemented and used to construct the water distribution network in an optimized way by [3] in 2003. In several various applications, it has been successfully used and gives satisfactory performance [7]. The original algorithm also reveals some drawbacks, such as non-uniform initial population, local and adaptive power limitations, and premature convergence [8].

It is evident from the literature that extensive surveys on certain algorithms inspired by nature, have been performed, such as PSO [9], Whale Optimization Algorithm (WOA) [10], Dragonfly Algorithm(DA) [11], and Backtracking Search optimization Algorithm (BSA) [12, 13]. Nonetheless, no recent survey or research work has been conducted on SFLA since 2014 as it is a very distinguished algorithm and has many applications and publications rely on it to provide a summary on the algorithm, its adjustments, its hybridizations, and its applications. After its initial journal publishing, SFLA has gained widespread recognition. It has subsequently achieved widespread acceptance in a variety of engineering and medical sectors, resulting in an enormous amount of published work, as shown in Figure (1). It demonstrates a sustained and increasing interest in the SFLA method and its variants, most notably since 2006 in terms of publications on its use, but also in terms of significant algorithm layout modifications since 2006. We plan to include a thorough overview of SFLA in this analysis in order to provide a valuable resource for researchers looking for inspiration for emerging technology in a variety of realistic applications. We would make every attempt to implement the majority of these developments.

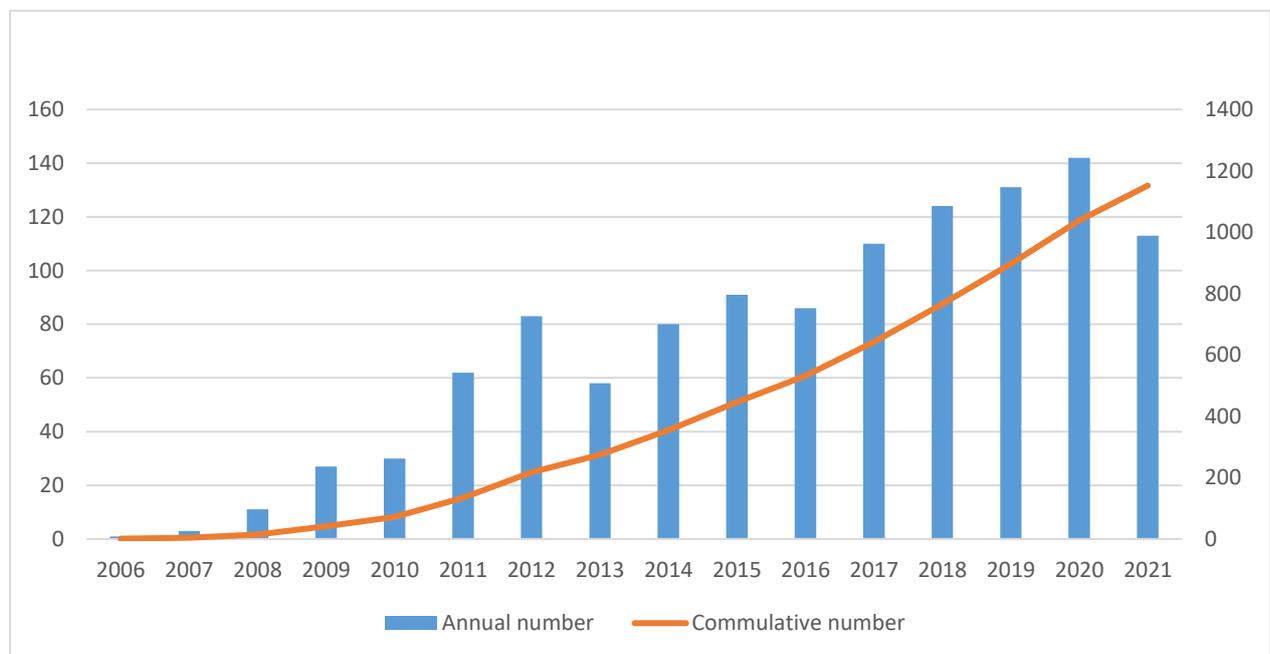

**Figure (1): Growing interest in the analysis and application of SFLA since 2006 (Source: Google Scholar, keyword search: "Shuffled Frog Leaping Algorithm", at 10: 42 Wednesday 3$^{rd}$ November 2021)**

The main reason behind conducting this review is to help researchers use it in different areas or propose modifications to it and hybridize it with other common algorithms. The purpose of this analysis consisted of several aspects: first, the most research developments and studies carried out on SFLA were highlighted, where hybridization models were used to combine SFLA with other techniques to improve the resulting algorithm's efficiency. Second, this research has concentrated of adjustment that have been applied to SFLA to enhance its power to look for the best solution. Third, much of the research work related to different applications applied to SFLA has been collected. In return, this research work would also open the way for researchers to introduce other improvements to the SFLA algorithm to match their various purposes in diverse fields.

The outline of this paper starts with summarizing the previous literature work on SFLA, followed by describing SFLA with its characteristics in Section 2. Additionally, various SFLA modifications are presented and explained and in Section 3. Section 4 is about the extensions of SFLA. Next, various applications of SFLA are presented in Section 5. Then, an operational framework of SLFA and its variants is proposed in Section 6 to analyze their uses on different cohorts of applications. Finally, after summarizing the enhancements and evaluating future research directions, we conclude the key achievements form this review article.

## 2. Related Work

This section outlines the existing review work on SFLA. This is the only survey on SFLA conducted by [7]. It was about a survey on the role of basic, modified and hybrid SFLA on optimization problems. The article contrasted prior studies on SFLA and its usefulness to the most widely used optimization algorithms. Referencing to the previous literature work, many efforts by previous researchers on SFLA denote the next generations of basic SFLA with diverse structures for modified SFLA or hybrid SFLA. As well, an attempt is made to highlight these structures, their enhancements and advantages. Additionally, this review paper discussed the top SFLA improvements for solving multi-objective optimization issues, including optimizing local and global discovery, preventing local optima, decreasing computational time, and improving the efficiency of the initial community. The calculated improvements in SFLA focused on statistical analysis of 89 reported articles and taking into account the most often used and successful changes made by a significant number of researchers. Finally, the quantitative validations address the SFLA as a versatile algorithm that outperformed other optimization algorithms in a variety of applications.

Our review study distinguishes itself from the earlier survey work in a variety of ways, as detailed below: (1) The previous study is out of date, having been published in 2014, and it did not concentrate on different types of applications, such as engineering, power and energy, and computing. This overview discusses the most current advancements in a variety of applications from 2006 to 2021. (2) The implementation of SFLA is described in detail, including a step-by-step approach and a practical example. (3) We also discuss the state-of-the-art reviews on SFLA and its derivatives. (4) An operational framework for SFLA is presented in order to examine its core implementation and extensions in real-world applications and to establish links between these versions. (5) It not only demonstrates the difficulties of SFLA, but also suggests possible solutions. We identified over 500 recent research papers by entering different versions of the keyword 'shuffled frog-leaping algorithm and SFLA' into the Google Scholar search box. Following a rigorous study of the gathered papers, we picked around 70 as the most significant in terms of SFLA and its variations on practical applications, which are addressed in this study.

## 3. The standard SFLA

As a meta-heuristic, the SFLA was suggested, and it focuses on the progression of memes between collaborating individuals and the information global exchange. Conventional mathematical models are based on population principles; thus, a group of agents is a population. The agent has the corresponding utility value, which controls the individual value. Time is separated into time loop steps, which are discreet. Figure (2) illustrates the SFLA flowchart. The algorithm is defined in the following subsections.

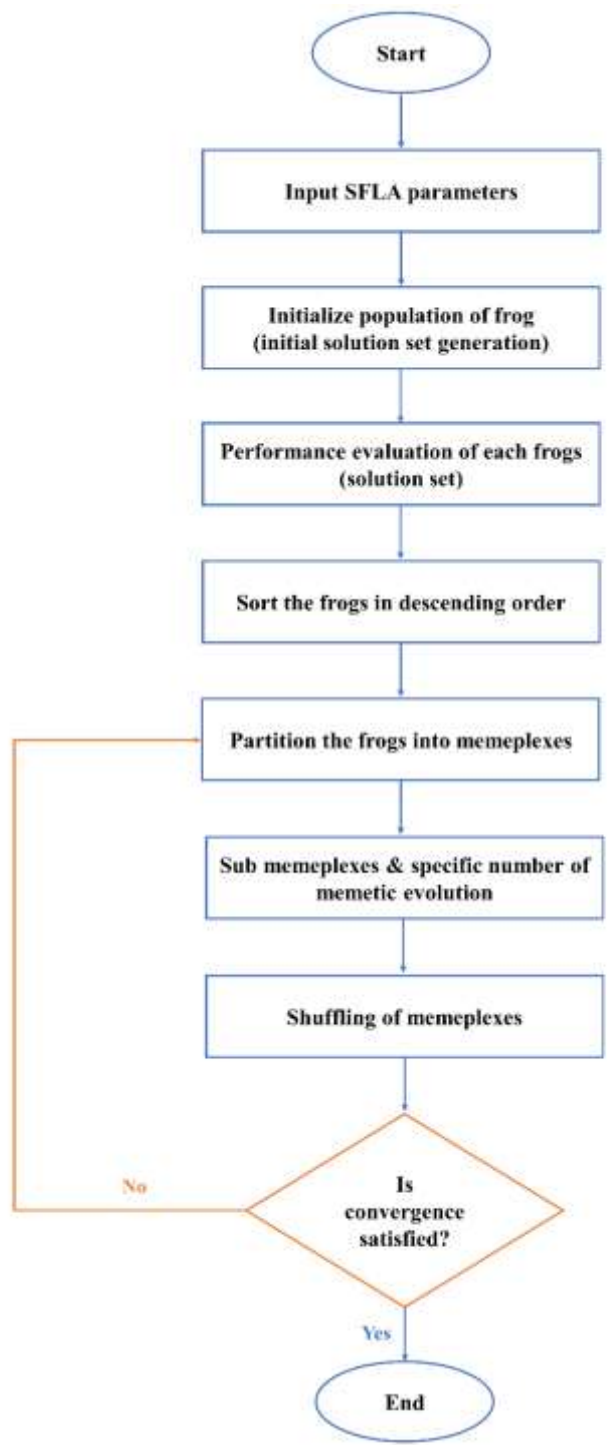

**Figure (2): The flowchart of SFLA**

## 3.1. The Standard SFLA

The population comprises a swarm of frogs or solutions, split up into subgroups identified as memeplexes [14]. Each frog consists of memo types (decision variable parameter). The population of frogs is shown as $U_1, U_2, \ldots, U_F$ and the frog $i$ is presented as a decision vector variable such as $U_i = \{U_i^1, U_i^2, \ldots, U_i^d\}$, $d$ signifies the number of decision variables in a frog. Memeplexes each conducting a local search. After a certain number of evolution stages, ideas are transferred across memeplexes through a shuffling mechanism. Both the local search and shuffle processes are continued until the convergence criteria stated are satisfied.

In the two phases, the SFLA meta-heuristic technique is outlined [3, 15]:

1) A phase of the global exploration consists of the following steps:
   a. <u>Step One:</u> Initialization of Population
      i. specify the memeplexes number; $m$.
      ii. specify the frog's number in each memeplex; $n$
      iii. Generate the population; $F = m \times n$
   b. <u>Step Two:</u> Generation of virtual population.
      i. Generate sample $F$; virtual frogs $U_1, U_2, \ldots, U_F$ in the achievable space $\Omega \subset \aleph^d$. $d$ signifies decision variables' number, for instance, memotype (s) number in a meme held by a frog.
      ii. Calculate the fitness function for each frog,
   c. <u>Step Three:</u> Frogs ranking.
      i. arrange the $F$ frogs in descending order based on their fitness function and place them in a list $X$; $X = \{U_i, f_{(i)}, i = 1, \ldots, F\}$, so as i=1 characterizes the best global frog ($P_x$), where $P_x = U_1$.
   d. <u>Step Four:</u> Divide frogs into memeplexes. For example, memeplex 1 gets rank 1, memeplex 2 gets rank 2, memeplex $m$ gets rank $m$, memeplex 1 gets rank $m+1$, this process continue until all frogs has been partitioned on the memeplexes.
   e. <u>Step Five:</u> Evolution of Memetic inside each memeplex.
      i. Entering the phase of local exploration.
   f. <u>Step Six:</u> Mix up memeplexes.
   g. <u>Step Seven:</u> Verify convergence.
      i. Terminate if criteria of convergence are met. If not, go back to step 3.

2) The phase of Local exploration: Each memeplex progression continues autonomously in Step Four of the global search *N* times. The algorithm returns to global shuffling exploration after the memeplexes have been progressed. The steps of the local exploration phase are described in detail for each memeplex in the local search.

   a. <u>Step One:</u> Set the counters for no. of memeplexes(*m*) and counting evolutionary steps N in each memeplex such that:
      i. *im=m+1*
      ii. *iN=N+1*
   b. <u>Step Two:</u> Form a sub-memeplex for each memeplex.
      i. The sub-memeplex suggested technique assigns greater weight to frogs with greater performance and fewer weight frogs to lower performance. Through a triangular probability distribution, the weights are allocated as in Equation (1) [3].

$$p_n = \frac{2(n+1-j)}{n(n+1)} \qquad \textit{Where j=1, 2,...,n} \qquad (1)$$

   Here, a submemeplex of array *Z* is formed by a random selection of *q* distinct frogs from n frogs in each memeplex. The frogs arranged in decreasing order such that (*iq=1*) is the frog's best position ($P_B$) $and$ ($iq = q$) is the frog's worst position ($P_W$). The idea of the submemeplex is exemplified in Figure (3).

   c. <u>Step Three:</u> Calculate step size S with Equation (2) [3] and enhance the position of the worst frog using Equation (3) [3].

*step size S =min {int [rand ($P_B$ - $P_W$)], $S_{max}$}*     *for a positive step*
*step size S =max {int [rand ($P_B$-$P_W$)], $S_{max}$}*     *for a negative step*     (2)

   *rand* signifies the random number between ranges from zero to one and $S_{max}$ signifies the maximum step size allowed by a frog to jump.

$$U_{(q)} = P_W + S \qquad (3)$$

   Where $U_{(q)}$ shows a new place in the next iteration (See Figure (4)).

   d. <u>Step Four:</u> if new $U_{(q)}$ > old $U_{(q)}$ then go to the local exploration step, otherwise, go to Step Five.
   e. <u>Step Five:</u> Select the global best frog position $P_X$.
      i. Find new step size S by:

*Step size S =min {int [rand ($P_X$ - $P_W$)], $S_{max}$}*     *for a positive step*
*Step size S    =max {int [rand ($P_X$-$P_W$)],$S_{max}$}*     *for a negative step*

    ii. Then, calculate the new position through Equation (3).
- f. <u>Step Six:</u>
    - i. If $U_{(q)}$ could not resolve inside viable space, then calculate the new performance, or else, move to Step Six.
    - ii. If the value of $f_{(q)}$ was better, then switch the old frog with the new one and move to Step Seven, otherwise move to Step Six.
- g. <u>Step Seven:</u> Generate a new frog randomly.
- h. <u>Step Eight:</u> upgrade the memeplex.
- i. <u>Step Nine:</u> Move to Step Two, *If (iN < N),*
- j. <u>Step Ten:</u> Move to Step 1, *If (im < m)*, otherwise, move back the global search to shuffle memeplexes.

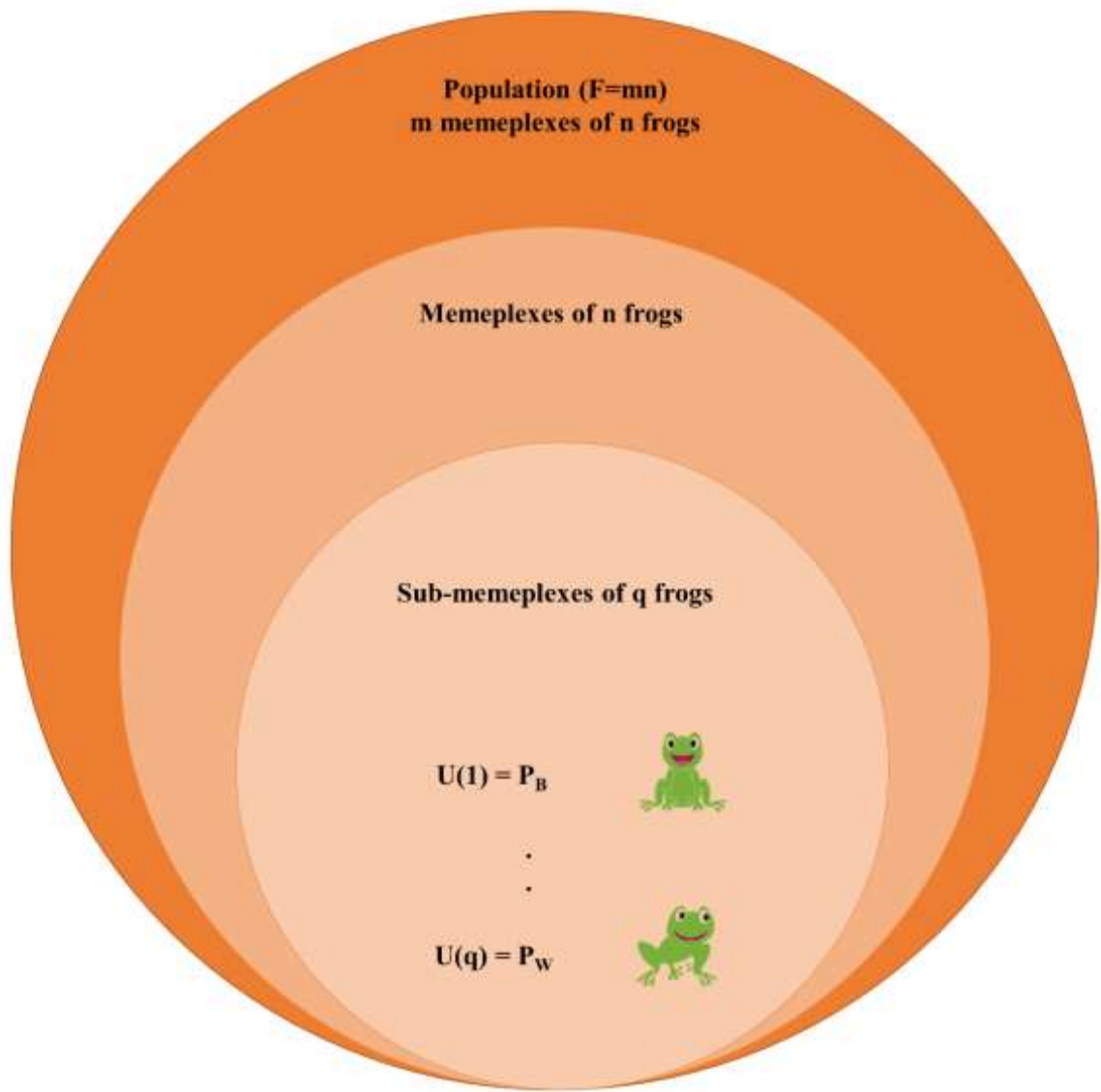

**Figure (3): An example of the idea of a sub memeplex. $P_B$ correspond to the position $U_{(1)}$ and $P_W$ correspond to the position $U_{(q)}$.**

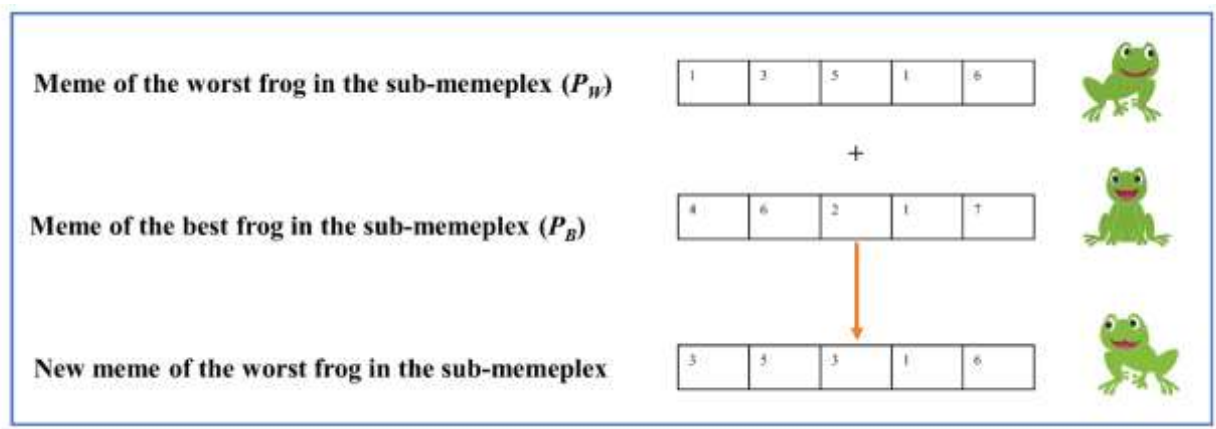

**Figure (4): An illustration of the memetic evolution in a sub-memeplex**

## 3.2. Hyperparameter Tuning for SFLA

Like all the other heuristic algorithms, selecting parameters is essential for the results of SFLA. SFLA consists of the following parameters:

1) *m,* signifies memeplexes number.
2) *n*, signifies frogs' number in a memeplex
3) *q signifies* frogs' number in a sub-memeplex
4) *N* signifies infection steps number between two consecutive shuffling in a memeplex.
5) $S_{max}$ signifies the size of the maximum step permitted in the course of an evolutionary step.

The hyperparameter tuning for SFLA and their impacts on algorithm output are shown in Table (1).

**Table (1): Hyperparameter Tuning for SFLA**

| Parameter | Description | Selection of small no. of the parameter leads to | Selection of large no. of the parameter leads to |
|---|---|---|---|
| F=(mn) | The most significant parameter is the number of F. | Increases locating the global optimum or near optimum probability. | Making it more computationally burdensome. |
| m, n, q | It is necessary to ensure that n is not too small | If too few frogs were presented in each memeplex, the use of the local memetic evolution strategy is lost.<br><br>The output reaction of the algorithm to *q* is that the exchange of information is slow if too few frogs are chosen in a submemeplex, resulting in greater solution times. | When several frogs were picked, the frogs were contaminated with undesirable thoughts that cause the phenomenon of censorship that aims to extend the searching time. |
| N | Any value bigger than 1 can be taken. | The memeplexes will be shuffled regularly, decreasing the exchange of ideas on a local scale. | Each memeplex can shrink into a local optimum. |
| $S_{max}$ | A limitation on regulating the global exploration capacity of the SFLA. | It decreases the global exploration, rendering the algorithm a local search. | The true optimum could be lacking, since it is not fine-tuned. |

## 4. Variants of SFLA

Due to the SFLA's success, several publications have been suggested, and many of them have been implemented in multiple implementations to improve the efficiency of the original SFLA [3]. Scholars were motivated to improve the performance of standard SFLA in light of its supposed shortcomings in terms of deficiency and convergence speed. Numerous authors have suggested revisions to the initial SFLA, although others have attempted adaptations to address a variety of problems. This segment addresses the evolution of SFLA variants and the present state of knowledge about them. These variations are divided into two groups: SFLA modifications (MSFLA) and SFLA hybridizations (HSFLA). The Figure (5) illustrates a high-level overview of the major SFLA variations.

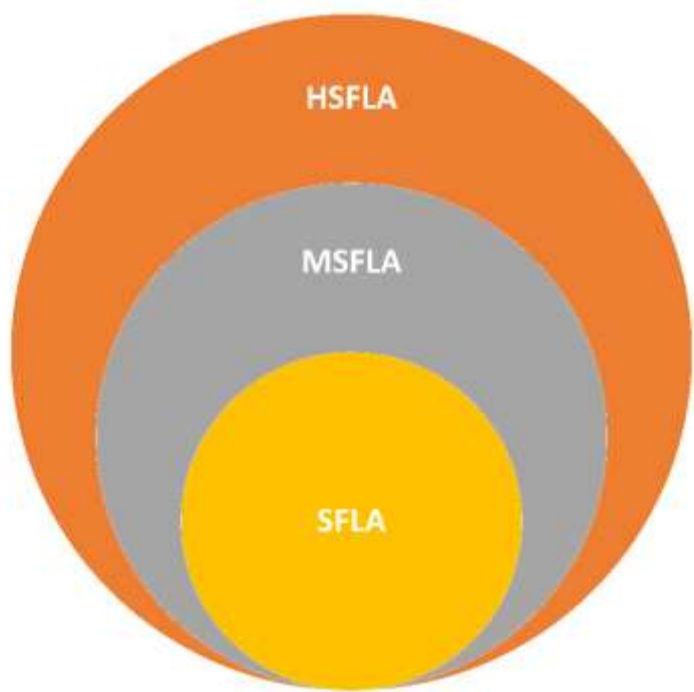

**Figure (5): The main variations of SFLA**

## 4.1. Modifications

Scholars have revised SFLA multiple times to improve its effectiveness and efficiency and to make it more fit for a specific use. This section examines nine different SFLA algorithms that have been modified. MSFLA, bilevel SFLA, two-phase SFLA, ISFLA, MCSFLA, MMSFLA, DSFLA, r-MQSFLA, and MGSFLA are the modified SFLA. Table (2) summarizes these algorithms. Some researchers in various areas have modified SFLA. Generally, the modifications are made in local search, global search, shuffling, memeplex, and leaping processes.

Table (2): The modified SFLA algorithms

| Algorithm | Reference, author(s) | Aim | Modification area | Methodology | Application | Results |
|---|---|---|---|---|---|---|
| MSFLA | [16], E. E. Elattar | Solving emission and economic dispatch (CHEED) issue by the use of green energy. | Global and local search processes in SFLA | Modifying the global and local search processes in SFLA | CHEED problem with renewable sources | The superiority of MSFLA over other methods was demonstrated. |
| | [17], M. Moazzami et al. | Solving the positioning and size of dispersed generating units and D-STATCO | Local search (Interactive updating of worst frog) | 1) Widening the frog leap space 2) Subgroup definitions | Distributed generators | MSFLA was more effective compared with the results achieved from genetic algorithm (GA). |
| | [18], M. H. Oboudi et al. | Providing a heuristic strategy for the purposeful islanding of microgrids | Problem solution space; bus voltage; load controllability; load priority; the ability to construct larger islands; and line capacity constraints; | Employing sub-memeplex | Microgrids | MSFLA is the ideal approach for TKP solution |
| | [19], D. Mora-Melia et al. | Water distribution network | Sub-memeplex parameter q | Optimization in parameter setting | Water distribution networks | The efficiency of MSFLA significantly improved. |
| | [20], A. Baziar et al. | Improving total searchability of the method and avoiding premature convergence | Two-phase modification (Lévy flight, population) | 1) Using random walk 2) Moving the average of the population toward the best solution. | Nonlinear and large scale problems | MSFLA showed superiority over the methods in the area. |
| | [21], T.-H. Huynh | Increasing the trajectory and duration of each frog | Local exploration | Proposing a new frog leaping rule | proportional-integral-derivative (PID) controller | The effectiveness of SFLA improved and premature convergence avoided. |
| | [22], E. Elbeltagi et al. | Overcoming local optimum and early convergence | Global search | Introducing a new parameter called a search acceleration factor (C) | Project management | The performance of SFLA improved. |
| Bilevel SFLA | [23], X. Duan et al. | Solving bi-level programming model | Combination of two basic models of SFLAs | Simulating the process of finding food for the frog population, which is separated into memeplexes | Emergency vehicle redistribution and dispatching | Bilevel SFLA was quick and efficient in resolving bilevel programming model. |
| Two-phase SFLA | [24], B. Naruka et al. | Overcoming the slow convergence of SFLA | Applying the elite Gaussian learning strategy in the global information exchange phase and updating the frog leaping rule and adding the learning attitude of the frogs | 1) Initialization of frogs using evenly random generated integers and opposition based learning (OBL) 2) The use of a scaling factor for new frog positions | Structural design problems | Increment in search range and enhancement in the population diversity were achieved. |
| ISFLA | [25], Z. Zhen et al. | Extending discrete optimization problem | Distributing the frogs into memeplexes | Combining local and global searches to enhances frog distribution randomness and variety in continuous space | Continuous optimization problem | ISFLA was capable of overcoming the premature convergence and sluggish convergence speed issues, as well as achieving high optimization accuracy. |
| | [26], J. Siahbalaee et al. | Relocating and reconfiguring distributed generators | Repositing the frogs towards a better position | Repositioning the frogs | Distributed generators | The proposed method had an advantage over previous experiments. |
| | [27], H.-P. Hsu and T.-L. Chiang | Examining complex and continuous berth allocation problem (DCBAP) | ISFLA has four features: numerous frog groups, shuffling method, discrete operators with self-adaptive leap, and self-adaptive mutation process | ISFLA possesses the following features: frog groups, shuffling method, discrete operators with self-adaptive jump, and self-adaptive | Berth allocation problem | ISFLA was capable to manage DCBAP at a container terminal. |
| | | | | mutation mechanism | | |
| | [28], T. Zhang et al. | Local backlight dimming | Initial solution generation and interval search solutions | Introducing cycle optimization in ISFLA | Local dimming technology | ISFLA delivers a better trade-off between power usage and picture quality than previous image parameter-based algorithms. |
| | [29], B. Hu et al. | Optimizing biomedical data | Chaos memory weight factor | Introducing an absolute balance group strategy, a chaos memory weight factor, and an adaptive transfer factor | Biomedical data | ISFLA improved both the detection of related subsets and classification accuracy. |
| MCSFLA | [30], H. Pu et al. | Optimizing unmanned aerial vehicle (UAV) | Mixing the concept of GA PSO | SFLA combination with genetic and social behavior-based particle swarm optimization | UAV flight controller | The dynamic reaction time of SFLA was very fast. |
| MMSFLA | [31], R. Azizipanah-Abarghooee et al. | Optimizing multi-objective power flow with flexible AC transmission systems (FACTS) | Mutation procedure | A new mutation with SFLA | Power flow with FACTS devices | MMSFLA was more effective compared to other algorithms. |
| DSFLA | [32], J. Cai et al. | Minimizing makespan in distributed scheduling problems | Modification in local search | Applying Dynamic search in each memeplex | Flow shop scheduling | It showed superiority over the methods in the area. |
| r-MQSFLA | [33], Y. Guo et al. | Solving an optimization problem of many-objective allocation of water resources for Inter-Basin water transfer IBWT | Global Search | A real-coded quantum computer and an external dynamic archive with SFLA | Water resources allocation | r-MQSFLA performed admirably in Jiangsu, China's Eastern Route of the South-to-North Water Transfer Project. |
| MGSFLA | [14], D. Lei and T. Wang | Solving distributed two-stage hybrid flow shop scheduling | Memeplex | SFLA with memeplex grouping | Flow shop scheduling | MGSFLA has most promising advantage over the other methods. |

## 4.2. Hybridizations

Hybridized SFLA is a term that refers to the combination of SFLA with another method. The purpose of hybridized SFLA is to use the strengths of both SFLA and the other algorithms while compensating for their deficiencies. This section discusses eight hybridized BSA algorithms, which are DESFLA, FSA-SFLA, Quantum SFLA, K-means SFLA, SFLA-SVM, SFLA-BFO, and PSFLA. Table (3) summarizes these hybridized algorithms of SFLA.

Table (3): The hybridized SFLA algorithms

| Proposed algorithm | Reference, author(s) | Aim | Hybridized with | Methodology | Application | Results |
|---|---|---|---|---|---|---|
| DESFLA | [34], L. Yan et al. | Tackling low accuracy and avoid local optimum | DE | Updating the strategy of DE for global searching ability | Optimum power management unit (PMU) | The reduction of the number of PMU configuration was achieved compared to DEPSO. |
| | [35], B. Naruka et al. | Improving searching capability | DE | Improving the searching capability of SFLA and maintaining the convergence of the population of frogs | Three different types of chemical engineering problems (linear and nonlinear) | DESFLA showed to be very efficient in solving computational optimization problems. |
| FSA-SFLA | [36], Z. Lu et al. | Finding optimal solution region quickly | Fish swarm algorithm (FSA) | Using the FSA's ability for global optimization, and SFLA's ability for local optimization | Data mining | FSA-SFLA was capable of successfully reducing attribute dimensions while maintaining classification capability. |
| Quantum SFLA | [37], L. Wang and Y. Gong | Achieving a balance between global and local search and to increase its speed | Quantum evolutionary algorithm (QEA) | Enhancing the population division, speeding up convergence, and advancing the success ratio that reaches the optimum | 0-1 knapsack problems | QBSFLA offered a number of benefits, including rapid convergence, a strong global search capability, and high stability. |
| | [38], W. Li et al. | Collaborative recommendation system | Quantum computing | Introducing quantum computing into SFLA, and using quantum movement in | Collaborative filtering (CF) | QBSFLA enhanced the predictive accuracy of rating scores. |
| | | | | the DQSFLA to search the local optima | | |
| K-means SFLA | [39], Z. Tang and K. Luo | Enhancing clustering results | K-means | Employing chaotic local search, and introducing a searching mechanism for updating frog's position | UCI datasets (Iris, Wine, Balance-scale, and Housing) | K-means had better accuracy. |
| SFLA-SVM | [9], W. Li et al. | Forecasting wind photovoltaic battery power | SVM | Burdening blindness and randomness in parameter selection of SVM | Wind-photovoltaic-battery generation system | SFLA-SVM had higher optimization capability. |
| SFLA-BFO | [40], Y. Li and Z. Yan | Reliability analysis | BFO | - Using a strategy of random grouping to maintaining the diversity of the population<br>- Using the update strategy of Levy flight to increase the global search ability<br>- Using the migration operation approach to escape the local optimums | System reliability model | It was capable of obtaining the optimal solution while maximizing the system's dependability. |
| PSFLA | [41], X. Nie and H. Nie | Avoid falling into the partial optimal solution, and to improve the convergence speed | PSO | Proposing a new MPPT strategy of the PSFLA combined with recursive least square filtering | Maximum power point tracking (MPPT) | It was capable of successfully suppressing measurement noise effects. |
| SFLA-GA | [42], Ibrahim GJ et al. | Optimizing mobile cloud service composition | GA | Embedding GA with SFLA to encode the individuals for discrete problems | Mobile cloud components | It was achieved the improvement in the practicability of the service with bare minimum power utilization, response time, and cost for mobile cloud |

|  |  |  |  |  |  | components according to some other algorithms. |
|---|---|---|---|---|---|---|

## 5. Key Applications of SFLA

Many publications have been implemented SFLA for real-life applications because of its simplicity in implementation and fast convergence. The key uses of SFLA are in the following domains: water resources management, computing, power generation and energy management, and engineering.

### 5.1. Environments and Water Resources

One of the areas that SFLA used is water resources management. Table (4) briefs the main applications of SFLA in water resources management.

Table (4): Summary of SFLA applications in water resources management

| Reference, author(s) | Year | Algorithm | Purpose |
|---|---|---|---|
| [33], Y. Guo et al. | 2020 | MSFLA | Inter-basin water transfers |
| [43], G. Liu et al. | 2019 | MSFLA | Solving the optimization problem of Raman Fiber Amplifier (RFA)of multi-pumped design |
| [44], T. K. Sharma and D. Prakash | 2019 | SFLA | Optimization of air pollution emission |
| [45], D. Prakash et al. | 2017 | SFLA | Air pollution control |
| [19], D. Mora-Melia et al. | 2016 | MSFLA | Water distribution network optimization problems |

### 5.2. Computing

Another uses of SFLA is computing. These applications are summarized in Table (5).

Table (5): Summary of SFLA applications in computing

| Reference, author(s) | Year | Algorithm cohort | Purpose |
|---|---|---|---|
| [46], B. Boroumand et al. | 2021 | MSFLA | The mapping process for Network on Chip (NOC). |
| [47], M. Mohammadhosseini et al. | 2020 | SFLA | Finding the proper route for the sensor to send information to the sink or coordinator. |
| [48], R. K. Khadanga et al. | 2020 | HSFLA | PID controller for affective load frequency control. |

17 | P a g e

| [49], M. Karpagam et al. | 2020 | MSFLA | Scheduling |
| [26], Z. Zhen et al. | 2020 | MSFLA | |
| [50], S. Kayalvili and M. Selvam | 2019 | HSFLA | |
| [51], D. Lei and X. Guo | 2016 | SFLA | |
| [52], M. Karakoyun | 2019 | HSFLA | Traveling salesman problem (TSP). |
| [8], X. Luo et al. | 2008 | SFLA | |
| [53], L. Wang et al. | 2019 | HSFLA | Solving the service discovery and combinatorial optimization of outsourcing resources (COOR) problem |
| [54], M.-L. Pérez-Delgado | 2019 | SFLA | Quantizing color image. |
| [55], N. Yuvaraj and A. Sabari | 2017 | SFLA | Feature selection technique for Tweeter Sentiment classification. |
| [29], B. Hu et al. | 2016 | MSFLA | Optimizing high-dimensional biomedical data by feature selection |

### 5.3. Power and Energy

SFLA was mostly used in power generation and energy management. Table (6) presents the key applications of SLFA used in energy and power generation.

Table (6): Summary of SFLA applications in power generation and energy management

| Reference, author(s) | Year | Algorithmic cohort | Purpose |
|---|---|---|---|
| [56], A. Amudha | 2014 | SFLA | Positioning of Unified Power Flow Controller in power line transmitter. |
| [57], M. E. Mosayebian et al. | 2016 | SFLA | Managing optimal operational micro-grid. |
| [18], M. H. Oboudi et al. | 2016 | MSFLA | Intentional islanding control of microgrids. |
| [41], X. Nie and H. Nie | 2017 | MSFLA | Control strategy of maximum power point tracking (MPPT) of PV. |
| [40], Y. Li and Z. Yan | 2018 | MSFLA | System reliability analysis. |
| [58], M. Gitizadeh et al. | 2013 | HSFLA | Distribution of generators |
| [14], D. Lei and T. Wang | 2019 | MSFLA | Flow-shop scheduling |
| [26], J. Siahbalaee et al. | 2019 | MSFLA | Power network distributions |

| [17], M. Moazzami et al. | 2017 | MSFLA | Sizing and locating distributed generation sources. |
| [16], E. E. Elattar | 2019 | MSFLA | Solving the combined heat, emission, and economic dispatch (CHEED) with wind and solar power problem |
| [59], A. S. Reddy and K. Vaisakh | 2013 | HSFLA | Economic load dispatch optimization problem |
| [60], R. Sridhar et al. | 2017 | SFLA | Maximum power point tracking (MPPT). |
| [61], S. Nagaraju et al. | 2019 | HSFLA | Combined Heat and Power Economic Dispatch problem (CHPED) |
| [62], P. Hu et al. | 2019 | HSFLA | Optimization of low impact development (LID) |
| [63], H. Chamandoust et al. | 2020 | SFLA | Tri-objective optimal performance of a smart hybrid energy system (SHES) |
| [64], D. Lei et al. | 2017 | SFLA | Solving a flexible job shop problem (FJSP) for energy consumption. |
| [65], A. Siadatan et al. | 2019 | HSFLA | Providing a fundamental modulation for multilevel inverter. |

## 5.4. Engineering Problems

SLFA used in several engineering areas. Table (7) lists the use of SFLA in different engineering domains.

**Table (7): Summary of SFLA applications in engineering**

| Reference, author(s) | Year | Algorithmic cohort | Purpose | Area |
|---|---|---|---|---|
| [23], X. Duan et al. | 2015 | MSFLA | Dispatching and redistribution of emergency vehicles on the highway network | Engineering |

| [66], T. Liu et al. | 2018 | HSFLA | Raman fiber amplifier design | Optical engineering |
|---|---|---|---|---|
| [67], A. Kaveh et al. | 2019 | HSFLA | Finding an optimal solution region rapidly in designing trusses | Civil engineering |
| [68], N. R. Babu et al. | 2020 | HSFLA | Optimizing TIDN controller values | Engineering |
| [69], A. Ghaemi and B. Arasteh | 2020 | MSFLA | Generating structural test data for automatic test data generation in software engineering | Software engineering |
| [70], F. Jiang et al. | 2020 | HSFLA | Inversion of 2D electrical resistivity problem | Engineering |
| [14], D. Lei and T. Wang | 2019 | HSFLA | Investigating distributed two-stage hybrid flow shop scheduling problem (DTHFSP) | Engineering |
| [27], H.-P. Hsu and T.-L. Chiang | 2019 | MSFLA | Organizing dynamic and continuous berth allocation problem (DCBAP) in which both arrived and incoming ships | Engineering management |
| [64], D. Lei et al. | 2017 | SFLA | Solving a flexible job shop problem (FJSP) for energy consumption | Engineering |
| [71], X. Tao et al. | 2019 | MSFLA | Power load-constrained time-cost-resource trade-off (TCRTO). | Civil engineering |

## 6. Operational framework of SFLA

As mentioned in the previous sections before, SFLA and its extension have been utilized in a variety of applications. Each SFLA variation application requires a unique process, although several SFLA variations may be used to address a given kind of issue, or several issues may be addressed with the same kind of SFLA variation. As a result, we present an operational framework for SFLA variations, which serves as a concise description of their applications and methods in environments and water resources, computing, power and energy, and engineering problems. Figure (6) illustrates the proposed framework. The main extensions of SFLA are HSFLA and HSFLA. The standard SFLA and its enhanced versions have been used to a variety of real-world applications. The standard SFLA has been deployed on nine types of different application problems. Similarly, MHSFA has been applied on nine different real-world applications. On the other hand, HSFLA was the most used one on different practical applications. It has been exploited on ten variety of practical problems. Meanwhile, the SFLA and its extensions have been used for the same problem in some cases. For instance, all the three mentioned algorithms have been utilized for flexible job/flow shop scheduling problem and power distribution grid and controller.

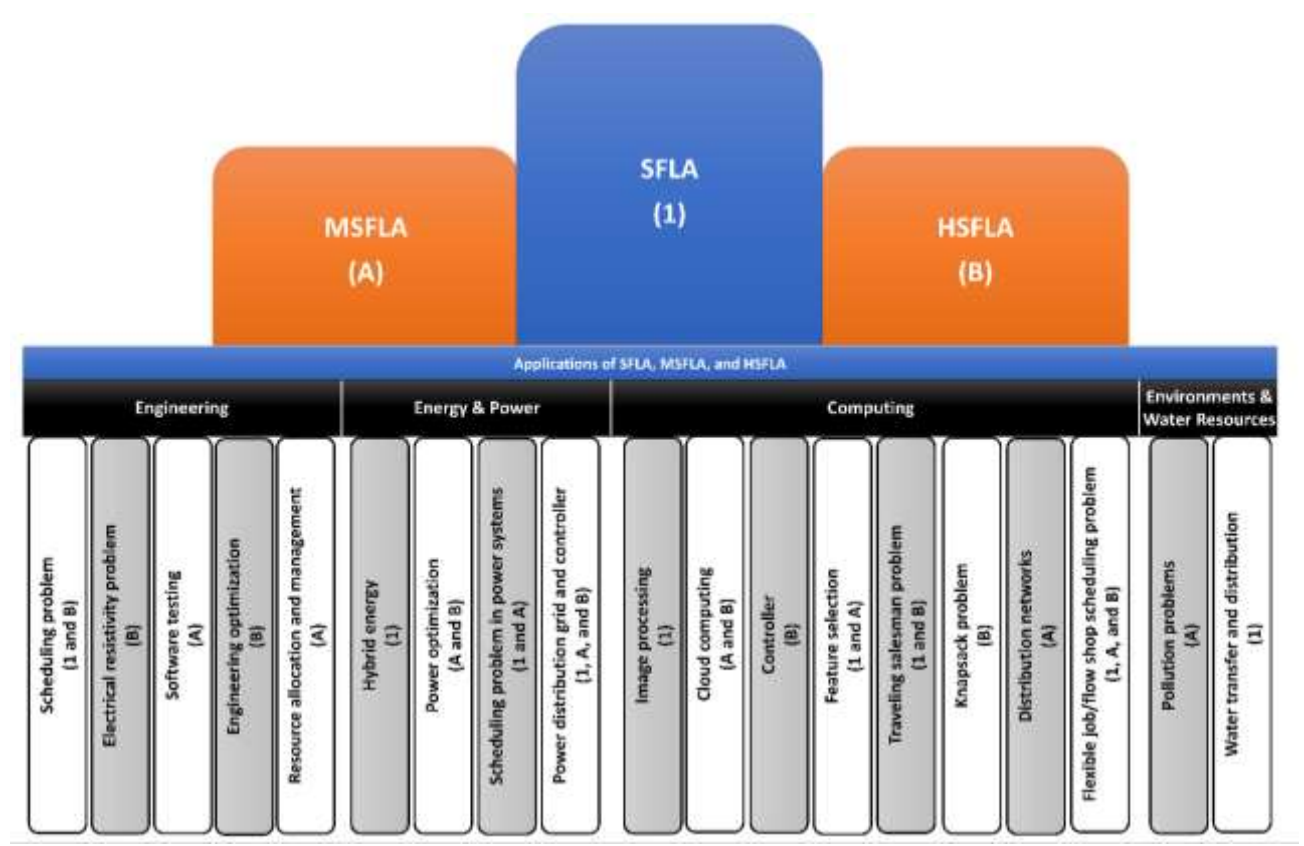

Figure (6): The proposed operational framework of SFLA

## 7. Conclusion and Prospects

In this review, a comprehensive survey was conducted, which can be used as a useful source by the scholars who desire to add an impact in the algorithm or apply it to solve real-world applications. In this work, we mentioned the most relevant and recent works applied in various areas, such as modifications done on the different parts of the SFLA. Also, the hybridization of the SFLA with other algorithms, and then, the applications of the algorithm were discussed. Finally, an operational framework for analyzing the usage of SLFA and its variations presented.

It can be seen from the statistics that the algorithm was in several real-world problems due to its simple implementation. Besides, it provides a solution to the complex problem as well due to its fast convergence. The performance of the algorithm was evaluated against other competitive algorithms, mostly with PSO and GA. In general, the SFLA has shown competitive results against both algorithms. From the modifications and hybridizations, which were done in the area, we can conclude that the SFLA has several aspects for researchers to work on, such as memeplex grouping and shuffling mechanism, parameter selection, virtual frog population creation, partitioning of frogs through the memeplexes, best and worst frog selection, and worst frog position updating formula. These limitations were overcome in local and global explorations for different proposes as preventing the algorithm from trapping into local optima and increasing the convergence rate.

Although SFLA and its variants were implemented to solve many complex problems, still some researchers claim some limitations in several aspects.

- Narrow local search space [21], since the worst frog can never jump over the best frog, which leads the algorithm to trap in local optima, reduces the convergence speed and precipitate convergence;
- Most of the publications claim about the slow convergence of the algorithm as well as trapping into local optima, and premature convergence [17, 20, 22, 24];
- Unbalanced memeplex performance [25];
- Long execution time [28].
- Because of its simplicity in implementation and the fewer parameters, many works have applied the SFLA on different applications and besides, the modifications, as well as hybridizations of the algorithm have been a hot spot for the scholars. But still, there is room for others to work in improving the algorithm as it can be hybridized with most recent swarm algorithms, such as backtracking search optimization algorithm [12, 13], the variants of evolutionary clustering algorithm star [72–75], chaotic sine cosine firefly algorithm [76], and hybrid artificial intelligence algorithms [77]. Furthermore, HS can be applied to more complex and real-world applications to explore more deeply the advantages and drawbacks of the algorithm or improve its efficiencies, such as engineering application problems [76], laboratory management [78], e-organization and e-government services [79], online analytical processing [80], web science [81], the Semantic Web ontology learning [82], chronic wound image processing [83], signal detection processing [84], and concept drift detection in big social data [85].

## Acknowledgements

Some special thanks go to Kurdistan Institution for Strategic Studies and Scientific Research and the University of Kurdistan Hewler for its help and willingness to conduct this review.

## Compliance with Ethical Standards

**Conflict of interest:** None.

**Funding:** No funding is applicable.